\documentclass[10pt,twocolumn,letterpaper]{article}

\usepackage[pagenumbers]{cvpr} 

\usepackage{graphicx}
\usepackage{amsmath}
\usepackage{amssymb}
\usepackage{booktabs}

\usepackage[pagebackref,breaklinks,colorlinks]{hyperref}

\usepackage[capitalize]{cleveref}
\crefname{section}{Sec.}{Secs.}
\Crefname{section}{Section}{Sections}
\Crefname{table}{Table}{Tables}
\crefname{table}{Tab.}{Tabs.}

\def\cvprPaperID{*****} 
\def\confName{CVPR}
\def\confYear{2022}

\begin{document}

\title{Vision Transformer for Action Units Detection}

\author{Vu Ngoc Tu\\
Chonnam National University\\
Gwangju, South Korea\\
{\tt\small tu369@jnu.ac.kr}
\and
Huynh Van Thong\\
Chonnam National University\\
Gwangju, South Korea\\
{\tt\small vthuynh@jnu.ac.kr}
\and
 Soo-Hyung Kim*\\
Chonnam National University\\
Gwangju, South Korea\\
{\tt\small shkim@jnu.ac.kr}
}
\maketitle

\begin{abstract}
Facial Action Units detection (FAUs) represents a fine-grained classification problem that involves identifying different units on the human face, as defined by the Facial Action Coding System. In this paper, we present a simple yet efficient Vision Transformer-based approach for addressing the task of Action Units (AU) detection in the context of Affective Behavior Analysis in-the-wild (ABAW) competition. We employ the Video Vision Transformer(ViViT) Network to capture the temporal facial change in the video. Besides, to reduce massive size of the Vision Transformers model, we replace the ViViT feature extraction layers with the CNN backbone (Regnet). Our model outperform the baseline model of ABAW 2023 challenge \cite{kollias2023abaw}, with a notable 14\% difference in result. Furthermore, the achieved results are comparable to those of the top three teams in the previous ABAW 2022 challenge. 

\end{abstract}

\section{Introduction}
\label{sec:intro}

Affective computing is a foundation field in Artificial Intelligence that aims to enable machines to recognize, interpret, and respond to human emotions. Recent advances in deep learning and computer vision techniques have enabled significant breakthroughs in the field, but several challenges remain unsolved. In particular, Facial Affect Analysis in the Wild emerge as a notable challenge in recent years. This task plays a crucial role in applications such as Human-Machine Interaction and serves as an initial step for many systems. As such, the Affective Behavior Analysis in the Wild (ABAW) competition \cite{kollias2022abaw,kollias2023abaw,kollias2023abaw2,kollias2021distribution,kollias2021analysing,kollias2021affect,kollias2020analysing,kollias2019expression,kollias2019face,kollias2019deep,zafeiriou2017aff} was organized to address these challenges. Since the first Workshop \cite{zafeiriou2017aff}, ABAW has become an important platform for researchers to benchmark their approaches and collaborate on solving Affective Computing problems.

The competition comprises three tasks focusing on detecting and recognizing three commonly-used presentations in affect analysis: Expression, Facial Action Units, and Valence-Arousal. 
The competition involves three tasks that focus on different affect presentations of affect analysis. The Facial Action Units (AU) detection task utilizes the Unit defined by the Facial Action Coding System (FACS) \cite{ekman1978facial} to capture and interpret facial muscle movements associated with different expressions. The Expression Recognition task, on the other hand, employs categorical and explicit definitions to represent human expressions. Finally, the Valence and Arousal estimation task uses continuous values to describe human emotional states, providing a more nuanced and comprehensive approach to affect the analysis. In this research, we particularly focus on the Action Unit detection task, which is the Multi-labels (12 labels) classification.

The Transformer architecture \cite{vaswani2017attention} has gained widespread popularity as a model of choice in the field of Deep Learning. Its successor, the Attention-based model, has emerged as the state-of-the-art approach not only for Natural Language Processing (NLP) tasks but also for achieving significant performance in various Computer Vision problems, especially in classification tasks \cite{dosovitskiy2021an}. For that reason, in this study, we present a Video Vision Transformer \cite{arnab2021vivit} based approach for the Action Units Detection task in the ABAW 2023 challenge.

Overall, the contribution of this paper can be summarized as following:

\begin{itemize}
    \item Instead of feeding the model raw video, we employ the CNN feature extraction model as the embedding module to get the presentation of video. This methods reduce the size of the model while still keeping the important information, hence help lightening the model.
    \item We utilize the ViVit model with Ensemble learning scheme for Facial Action Units model. The model outperform the baseline model and show competitive result comparing with the winner methods of last ABAW competition.
\end{itemize}

\begin{figure*}
  \centering
   \includegraphics[width=0.8\linewidth]{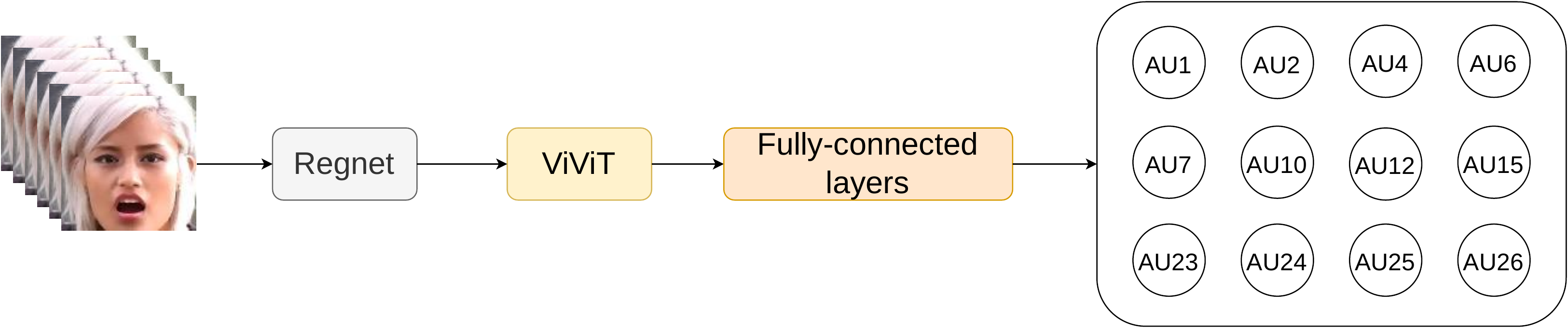}

   \caption{
   An overview of the action unit detection model.}
   \label{fig:onecol}
\end{figure*}

\section{Related Work}
\subsection{Facial Action Units Detection in the wild}

Facial affective computing has been a significant challenge in the field of computer vision since its early stage. During this period, popular approaches primarily focus on single-modal affective analysis and statical 2D data processing \cite{cowie2001emotion}. In the deep learning era, this problem have been extended to different manner such as  spontaneous facial expression \cite{mavadati2013disfa, zhang2014bp4d}, multi-modalities, 3D facial expression \cite{zhang2014bp4d} or automatic affect facial analysis \cite{martinez2017automatic}. DISFA \cite{mavadati2013disfa} is one of the first publicly available databases of spontaneous facial expression with well-annotatated Action Units intensity. One year later, Zhang et al introduce the first spotaneous 3D facial expression database \cite{zhang2014bp4d}. Until now, many Facial Affective Analysis researchs still uses these two dataset and attain remarkable result \cite{shao2021jaa, jacob2021facial}.

Although having been research for a long time, most Affective Computing research are done in the controlled environment. Therefore, the applicability of these research in real world is quite limited despite of having achieved good results. To address this problem, Zafeiriou et al \cite{zafeiriou2017aff} introduce the first dataset for analysing human affective Behavior in real world scenario.
The dataset can be access through the ABAW competition and is still updating each year.

\subsection{Vision Transformers in Facial Actions Units Detection}
Since the outstanding performance of ViT \cite{dosovitskiy2021an} and other NLP-influenced model in Image Classification and Video Classification, Transformer \cite{vaswani2017attention} has been adopted for various Computer Vision tasks. For Action Units Detection, "Facial Action Unit Detection With Transformers" \cite{jacob2021facial}
 is one of the first study using the Transformer. Following previous Region of Interest attention based methods such as JAA-Net \cite{shao2021jaa} or EAC-Net \cite{li2017eac}, the study use Transformer's Multi-head Attention Module as the ROI Attention module. While this methods achieve promising results on images, the size of model due to ROI Attention module may not be suitable for videos.
 
In the previous year's ABAW competition (CVPR 2022) \cite{kollias2022abaw}, a number of competitors employed Transformer. Specifically, among the top five teams ranked highest, three out of five teams utilized Transformer models as a core component of their model \cite{nguyen2022ensemble, wang2022multi, zhang2022transformer}. The winner team \cite{zhang2022transformer} use Transformer as an fusion module for their Multi-Modal architecture. On the other hand, the fourth places team \cite{wang2022multi} also use Multi-Modal scheme but Transformer is used for Feature Extraction purpose. While the third places team \cite{nguyen2022ensemble} treat each image feature from CNN extractor as a token and integrate Transformer to be the classification head. However, the Transformer model used in these methods are the original architecture which is not specifically suitable for video processing.

\section{Methodology}
Following the work of \cite{nguyen2022ensemble}, we construct a ViViT-based model \cite{arnab2021vivit} for Facial Action Units Detection problems. Our architecture comprises of two core modules: Feature extraction module and Classification module. Overall architecture is showed in Figure \ref{fig:onecol}.
\subsection{Feature Extraction}
For Feature Extraction module, due to good performance and small model size, we use RegNetY \cite{radosavovic2020designing} as the backbone. Proposed by Radosavovic et al in 2020, RegNetY is a type of simple and regular convolutional network. The RegNetY design space is defined by three primary parameters: depth, initial width, and slope, and generates a unique block width for each block in the network. Notably, RegNet models are constrained by a linear parameterization of block widths, meaning that the design space only includes models with one specific linear structure.
The RegNetY architecture is organized into multiple stages, each consisting of four blocks that collectively form the stem (start), body (main part), and head (end) of the network. Within the body section, multiple stages are defined, with each stage comprised of several blocks. It should be noted that RegNetY employs a single type of block throughout the network, specifically, the standard residual bottleneck block with group convolution. 

We adopt the Transfer Learning approach, leveraging a pre-trained RegNetY model that has been trained on the ImageNet dataset \cite{deng2009imagenet}. To be more specific, we use this pre-trained model as a backbone for training on the ABAW dataset. However, instead of freezing the entire backbone, we just partially unfreeze the last three blocks of the backbone while keeping the first block frozen. This allows us to fine-tune the pre-trained model for improved performance on our target task while still benefiting from the pre-existing knowledge gained during training on the large and diverse ImageNet dataset.

\subsection{Frames-wise Classification}
As mention earlier, we choose Video Vision Transformers \cite{arnab2021vivit} as the classification module. Inspired by ViT \cite{dosovitskiy2021an}, this model process on the a sequence of spatio-temporal tokens that extracts from the video. Transformer layers employ a distinct approach wherein all pairwise interactions among spatio-temporal tokens are modeled within each layer. As a result, the transformer layer is able to capture long-range dependencies across the entire video sequence from the initial layer itself. However, as the Multi-Headed Self Attention has quadratic complexity with respect to the number the tokens, we reduce the computational complexity by removing the first 4. We select the Factorized encoder variant of the ViViT model  based on its optimal trade-off between inference speed and accuracy, as compared to the other three variants.

From a extracted video embedding $V \in R^{B\times T\times E_l\times E_h\times E_w}$, with $E_l$, $E_h$, $E_w$ is the length, height and width of each frame embedding, we use Tubelet Embedding of ViVit to convert into sequence of token. Then the token will be fed into Transformer layers comprises of Multi-Headed Self-Attention (MSA) \cite{vaswani2017attention}, layer normalisation (LN) and MLP blocks. Each element is formalised in follow equation:
\begin{equation}
  y^l = MSA(LN(z^l)) + z^l
  \label{eq:important}
\end{equation}
\begin{equation}
  z^{ l+1} = MLP(LN(y^l)) + y^l
  \label{eq:important}
\end{equation}

\section{Experiments and results}
\subsection{Dataset}

The AU task contains 541 videos that include annotations in terms of 12 AUs, namely AU1, AU2, AU4, AU6, AU7, AU10, AU12, AU15, AU23, AU24, AU25, and AU26 of around 2.7M frames and contain 438 subjects, 268 of which are male and 170 female. The dataset have been annotated in a semi-automatic procedure
(that involves manual and automatic annotations).
\subsection{Experiments setup}
The networks were implemented with the PyTorch Deep Learning toolkit. We trained model by using SGD with learning rate of 0.9 combine and Cosine annealing warm restarts scheduler \cite{loshchilov2016sgdr}. The networks is optimized with Focal loss function \cite{lin2017focal}. 
The number of frames in each sequence length is set as 256 frames. For the ViViT model, we set number of Head in each Attention module is 8 heads and the hidden dimension of the transformer is 1024. Besides, we only keep the last 8 Transformer layers of ViVit and remove the rest. 
\subsection{Metrics}
According to challenge white paper \cite{kollias2023abaw}, macro F1 Score is the official evaluation criterion for Action Units detection task. Therefore,  the performance measure is calculated as  the average F1 Score across all 12 AUs:
\begin{equation}
P_AU = \frac{(\sum _{au} F_1^au)}{12} 
\end{equation}

\subsection{Results}
The results of our methods comparing with the baseline and other top-3 methods of previous competition is presented in \cref{tab:example}. According to the result, our methods significantly surpass the baseline methods with 14\% difference. On the other hands, our methods also achieves comparable performance with highest rank last year competitors.
Besides, we also report the results of the K-fold cross validation experiments with training set of ABAW challenge and evaluate on Validation set is showed in  \cref{tab:Fold}. As showing in the table, our model performance do not face overfitting problem in particular set and stabilize on the whole dataset.

\begin{table}
  \centering
  \begin{tabular}{@{}lc@{}lllc@ lc@{}}
    \toprule
    Methods & Val Set && Test Set  \\
    \midrule
    Baseline & 0.39 && \_\\
    Top 1 & 0.525  && 0.499\\
    Top 2 & 0.731 && 0.498\\
    Top 3 & 0.544 && 0.490\\
    Our methods & 0.5398&& \_\\
    \bottomrule
  \end{tabular}
  \caption{Compare with last year methods.}
  \label{tab:example}
\end{table}
\begin{table}
  \centering
  \begin{tabular}{@{}lc@{}}
    \toprule
    Fold & F1 Score  \\
    \midrule
    1 & 0.5211  \\
    2 & 0.5319 \\
    3 & 0.5277\\
    4 & 0.5332\\
    5 & 0.526 \\
    Val Set & 0.5398\\
    \bottomrule
  \end{tabular}
  \caption{Results. of our methods on Valdiation Set and on K-fold Cross-validation.}
  \label{tab:Fold}
\end{table}

\section{Conclusion}
In this paper, we present the Vision Transformer-based model for AU detection in the ABAW Competition. To reduce the computational burden and improve the effectiveness of our approach on small images, we propose using a CNN-based model for feature extraction instead of relying solely on the long Transformer backbone layers. Our method outperforms the baseline model and achieves competitive results compared to other methods used by last year's participants
\section*{Acknowledgements}
This work was supported by the National Research
Foundation of Korea (NRF) grant funded by the Korea government (MSIT) (NRF-2020R1A4A1019191) and Basic Science Research Program through the National Research Foundation of Korea (NRF) funded by the Ministry of Education (NRF2021R1I1A3A04036408).
{\small
\bibliographystyle{ieee_fullname}
\bibliography{PaperForReview}
}

\end{document}